\pdfoutput=1

\PassOptionsToPackage{table}{xcolor}
\documentclass[10pt,twocolumn,letterpaper]{article}
\usepackage[T1]{fontenc} %

\usepackage[pagenumbers]{cvpr}
\usepackage{booktabs}
\usepackage{multirow}
\usepackage{amssymb}

\newcommand{\ours}{TabAct}
\newcommand{\Ours}{TabAct}

\definecolor{cvprblue}{rgb}{0.21,0.49,0.74}
\usepackage[breaklinks,colorlinks,allcolors=cvprblue]{hyperref}

\title{Tabular Foundation Models Are\\
Surprisingly Effective In-Context Robot Learners}

\author{Zhen Xu \quad Hao-Run Cai \quad Heng-Zhuang Li \quad Si-Yang Liu \quad Jun-Peng Jiang \quad Han-Jia Ye\\
School of Artificial Intelligence, Nanjing University\\
National Key Laboratory for Novel Software Technology, Nanjing University
}

\begin{document}
\maketitle
\begin{abstract}
Can foundation models pretrained for generic
numerical prediction serve as in-context robot
learners? We investigate this question through
tabular foundation models (TFMs), which are
pretrained to infer input--output relationships from
labeled examples in context. We introduce \ours{}, which
casts visual imitation as numerical prediction: frozen visual features and
robot proprioception form the inputs, future action
chunks form the targets, and each demonstration time step
provides one labeled example that conditions a frozen
TFM for closed-loop control.
The predictor receives no robot-specific training
or target-task fine-tuning.
With one target-task demonstration available to each
method, \ours{} with TabICLv2 achieves the highest
average success among all evaluated methods on both
LIBERO and Meta-World, exceeding the strongest baseline
by 20.9 and 5.8 percentage points, respectively.
Additional demonstrations improve control without
updating model weights: with five demonstrations,
success reaches 94.2\% on LIBERO-Object. These results demonstrate that generic
numerical pretraining can provide an effective
foundation for in-context robot learning.
\end{abstract}

\begin{figure}[t]
    \centering
    \includegraphics[width=\columnwidth]{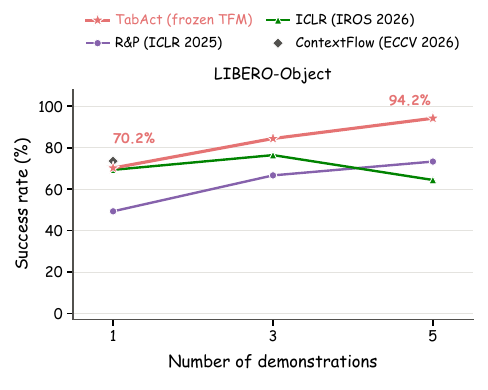}
    \caption{\textbf{A frozen TFM benefits from more demonstrations.}
    \Ours{} improves from 70.2\% success with one demonstration
    to 94.2\% with five on held-out LIBERO-Object tasks,
    with all pretrained weights frozen.
    Points show mean success across three demonstration groups.
    We compare against Retrieve and Play (R\&P),
    ICLR\protect\footnotemark, and ContextFlow;
    ContextFlow's released implementation supports only $K=1$.
    Section~\ref{sec:context_scaling} gives the protocol.}
    \label{fig:teaser}
\end{figure}
\footnotetext{ICLR denotes the method of Nguyen et al.~\cite{nguyen2026iclr}, not the conference.}

\begin{figure*}[t]
    \centering
    \includegraphics[width=\textwidth]{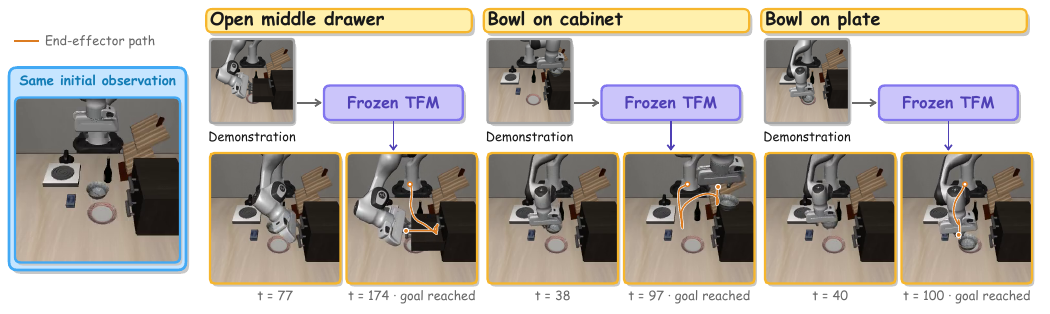}
    \caption{\textbf{Different demonstrations specify different goals from the same initial observation.}
    Three selected successful one-shot LIBERO-Goal rollouts use the same frozen
    TabICLv2 weights. Demonstration images are representative frames;
    orange curves show recorded end-effector paths projected into the camera view.
    Table~\ref{tab:goal_switch} reports aggregate results of the controlled
    goal-switching evaluation (Sec.~\ref{sec:context_behavior}).}
    \label{fig:goal_switch}
\end{figure*}

\section{Introduction}
\label{sec:intro}

Providing a few demonstrations is a natural way to teach
a robot a new behavior. In-context learning (ICL) makes
this possible without task-specific parameter updates,
using demonstrations as context for action prediction at
deployment. Robot-trained in-context policies acquire this
ability through training on robot demonstrations and
trajectories~\cite{duan2017oneshot,james2018tecn,dasari2021transformers,fu2024icrt,nguyen2026iclr,contextflow,vosylius2025instant,sridhar2025ricl}.
Language and multimodal foundation models offer another
route, drawing on in-context capabilities acquired through
general-purpose pretraining~\cite{brown2020gpt3,dipalo2024kat,yin2025roboprompt,roboicl}.
In this work, we investigate whether pretraining for generic
numerical prediction can provide this capability.

The connection starts with the structure of a demonstration.
Each time step pairs an observation with an action, so even
a single trajectory provides a small labeled dataset.
Acting from a new observation then requires predicting an
action from these examples. Once observations and actions
are represented numerically, this structure matches the
prediction interface of tabular foundation models (TFMs),
which are pretrained to infer input--output relationships
from labeled examples in context~\cite{tabpfn,tabiclv2}.
A demonstration can therefore supply the context for a TFM,
while the robot's current observation supplies the query.
\emph{Can a frozen predictor whose pretraining uses no robot
data use these numerical examples to support visual,
closed-loop robot control?}

We show that this is possible with \ours{}, a numerical
formulation of visual imitation for frozen TFMs.
\Ours{} converts demonstrations into compact input--output
examples: frozen visual features and robot proprioception
form the inputs, while short action chunks form the targets.
Projections fitted on the supplied demonstrations compress
visual features and future motion, reducing the dimensions
of the inputs and prediction targets. At deployment, the
TFM predicts an action chunk for the current observation,
conditioned on these demonstration examples. The robot
executes the predicted actions, obtains a new observation,
and repeats, placing the predictor directly in the control
loop. The visual encoder and TFM remain frozen throughout,
with no robot-specific policy training or target-task
fine-tuning.

We evaluate \ours{} in simulation on held-out manipulation
tasks from LIBERO~\cite{liu2023libero} and
Meta-World~\cite{yu2020metaworld}. With one target-task
demonstration available to each method, \ours{} with frozen
TabICLv2~\cite{tabiclv2} achieves 62.7\% average success
across three held-out LIBERO subsets and 49.3\% on
Meta-World. On both benchmark averages, it outperforms all
evaluated baselines, including robot-trained in-context
policies, retrieval-based imitation, and pretrained robot
policies fine-tuned on the target demonstration
(Table~\ref{tab:main_results}).

Beyond overall success, we examine how the frozen model
uses its demonstration context. Providing five demonstrations
instead of one raises average LIBERO success from 62.7\%
to 81.3\%, while all pretrained weights remain fixed
(Sec.~\ref{sec:context_behavior}). Figure~\ref{fig:teaser}
illustrates the gains on LIBERO-Object.
In a controlled goal-switching
evaluation, we hold the initial observation fixed and change
the supplied demonstration to specify different goals.
\Ours{} completes the demonstrated goal in 75.6\% of
rollouts, showing that the same frozen model can produce
different goal-directed behaviors through demonstration
conditioning (Fig.~\ref{fig:goal_switch};
Table~\ref{tab:goal_switch}). These results show that
demonstrations both specify the desired behavior and provide
additional examples that improve its execution.

We further examine whether effective control depends on a
particular predictor or visual representation. With
observation features, action targets, demonstration context,
and execution held fixed, replacing TabICLv2 with
nearest-neighbor, linear, or boosted-tree predictors reduces
average success on both benchmarks
(Table~\ref{tab:numerical_learners}). At the same time,
effective control extends across multiple TFMs and frozen
visual encoders (Tables~\ref{tab:tfm_generality}
and~\ref{tab:visual_encoders}), rather than being confined
to a single model pairing. Together, these findings support
generic numerical pretraining as a viable foundation for
in-context robot learning.

Our main contributions are:
\begin{itemize}
\item \textbf{Frozen TFMs as in-context robot learners.}
We show that TFMs pretrained on generic numerical tasks
without robot data can support visual, closed-loop robot
imitation through demonstration context.
\item \textbf{\ours{}, a numerical formulation of visual imitation.}
We represent demonstrations as input--output examples for a
frozen TFM, using frozen visual features and proprioception
as inputs and compact action chunks as targets.
\item \textbf{An empirical characterization of this capability.}
We evaluate one-shot control, demonstration-conditioned goal
switching, and improvement with additional demonstrations,
and examine predictor choice and generality across TFMs
and visual encoders.
\end{itemize}

\begin{figure*}[t]
    \centering
    \includegraphics[width=\textwidth]{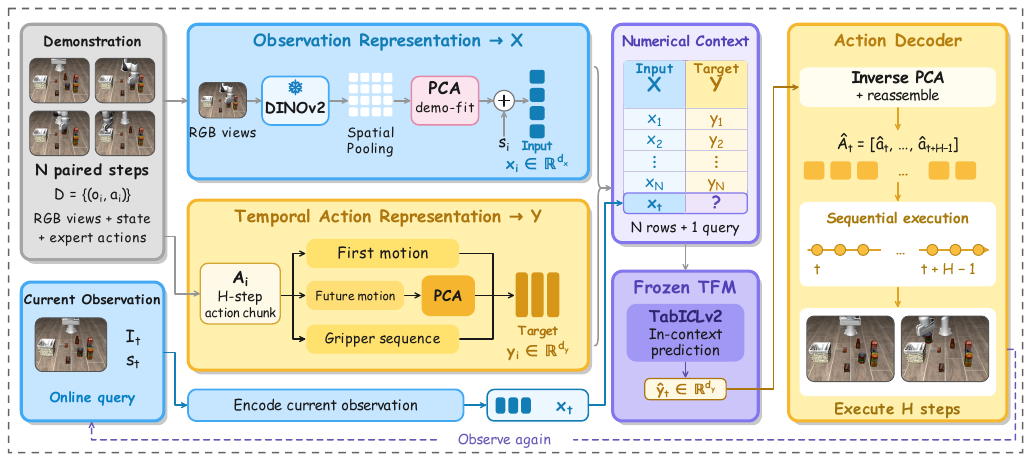}
    \caption{
        \textbf{Overview of \ours{}.}
        Expert demonstrations are converted into numerical
        input--output pairs using frozen visual features,
        demonstration-fitted PCA, and compact action chunks.
        These examples condition a frozen TFM (TabICLv2)
        to predict actions for the current observation.
        Predicted targets are reconstructed into action
        chunks for closed-loop control, shown with equal prediction
        horizon $H$ and execution horizon $E$ ($E=H$).
    }
    \label{fig:pipeline}
\end{figure*}

\section{Related Work}
\label{sec:related_work}

\paragraph{Robot-trained in-context imitation.}
Early one-shot imitation methods learn demonstration-conditioned
policies across robot tasks~\cite{duan2017oneshot,james2018tecn,huang2019ntg}.
Recent approaches use autoregressive action prediction
(ICRT~\cite{fu2024icrt}), visual reasoning traces
(ICLR~\cite{nguyen2026iclr}), or conditional flow matching
(ContextFlow~\cite{contextflow}). Instant Policy learns from
simulated pseudo-demonstrations~\cite{vosylius2025instant},
while RICL adds in-context adaptability to pretrained
vision-language-action models through additional training
on robot demonstrations~\cite{sridhar2025ricl}.
These methods acquire the ability to interpret demonstration
context through robot-policy training~\cite{jiang2023vima,patel2026bpp,mandi2022mosaic,jang2022bcz,jain2024vid2robot}. \Ours{} also conditions
behavior on demonstrations at deployment, but draws its
predictor's capability from generic numerical pretraining.

\paragraph{Language and multimodal models for robot ICL.}
Pretrained language and multimodal models offer another
route to in-context robot learning. A central design choice
is how to represent observations and actions for their
prediction interfaces. KAT~\cite{dipalo2024kat} converts
visual keypoints and action trajectories into text tokens,
while RoboPrompt~\cite{yin2025roboprompt} constructs textual
demonstrations from object poses and keyframe actions.
Multimodal approaches also combine demonstration context
with interaction feedback or
memory~\cite{roboicl,cheng2026gptpolicy,fang2026worldicl}.
\Ours{} shares this use of observation--action representations
to connect general-purpose pretraining to robot imitation.
We investigate numerical prediction as the source of this
capability, representing observations and action chunks as
labeled examples for a tabular foundation model.

\paragraph{Retrieval-based imitation.}
Retrieval-based imitation uses demonstrations as an action
memory~\cite{he2026alt,pfeifer2026darp,dipalo2024dinobot}. VINN~\cite{pari2022vinn} learns visual representations
for nearest-neighbor action prediction, while Retrieve and
Play (R\&P) retrieves actions directly from demonstration
examples. R\&P is the retrieval component of
REGENT~\cite{sridhar2025regent}; the full system combines
retrieval with a policy trained on queries and retrieved
examples. \Ours{} likewise treats demonstrations as an
inference-time dataset, using a frozen numerical predictor
to infer actions from the examples. This shared structure
motivates controlled comparisons between numerical
prediction and nearest-neighbor retrieval.

\paragraph{Tabular foundation models.}
Numerical in-context learning studies how models predict
outputs for new queries from labeled
examples~\cite{garg2022functions,akyurek2023algorithm,vonoswald2023gradient}. Prior-data fitted
networks~\cite{muller2022pfn} acquire this capability by
pretraining on tasks sampled from a prior to approximate
Bayesian prediction. TabPFN~\cite{hollmann2023tabpfn,tabpfn}
and TabICL~\cite{qu2025tabicl,tabiclv2} develop this approach
for tabular data, while TabDPT~\cite{ma2025tabdpt} studies
real-data pretraining with retrieval-based context.
In robotics, prior work uses a pretrained TFM as a return
surrogate to screen candidate policies during
optimization~\cite{ou2026tfms3}. We study a different role:
the TFM directly predicts action chunks from visual
demonstration examples and the current observation.
These predictions are executed in closed loop, determining
the observations encountered next~\cite{ross2011dagger}.

\section{Numerical In-Context Robot Learning}
\label{sec:method}

\subsection{From Demonstrations to Numerical Prediction}
\label{sec:method_motivation}

A demonstration contains many observation--action examples,
providing a task-specific dataset for predicting how the
robot should act. Retrieval-based imitation uses such
examples by matching the current observation to demonstration
observations and reusing the associated actions
\cite{pari2022vinn,sridhar2025regent}.
We retain the example-conditioned prediction structure and
use a frozen TFM as the predictor: labeled numerical examples
form its context, and the current observation serves as a
query (Fig.~\ref{fig:pipeline}). The model's predictive
capability comes from generic numerical pretraining.

A TFM takes a table of labeled rows as context and
predicts the label of a query row. Two properties of
these models shape our formulation. They are pretrained
on tables of modest width, so inputs must be compact
feature vectors rather than images or full feature maps.
Standard TFMs also predict one scalar target per query,
so the cost of predicting an action chunk grows with its
number of scalar targets. We therefore compress visual
features and future motion using demonstration-fitted
projections, while leaving the immediate action unchanged
(Sec.~\ref{sec:input_output_interface}). Because the
resulting examples are plain input--output pairs, other
predictors, including nearest-neighbor retrieval, can use
the same context, which enables the controlled predictor
comparisons in Sec.~\ref{sec:main_results}.

\paragraph{Problem setup.}
Given a target task $\tau$, the robot receives $K$
expert demonstrations
$\mathcal{D}_{\tau}=\{D_{\tau}^{(k)}\}_{k=1}^{K}$,
where demonstration $k$ contains $T_k$ observation--action pairs,
$D_{\tau}^{(k)}
=\{(o_t^{(k)},a_t^{(k)})\}_{t=1}^{T_k}$.
An observation $o_t=(I_t,s_t)$ contains RGB images
$I_t=\{I_t^{(c)}\}_{c=1}^{C}$ from $C$ camera views
and robot proprioception $s_t$.
An action $a_t=(u_t,g_t)$ consists of continuous
end-effector motion $u_t$ and a gripper command $g_t$.
The objective is to execute the demonstrated task under
new initial configurations, without updating the
pretrained model parameters.

\subsection{Constructing Numerical Inputs and Targets}
\label{sec:input_output_interface}

\paragraph{Numerical observations.}
\label{sec:numerical_context}

To represent images as numerical inputs, we use a frozen
pretrained visual encoder $\phi$, following the use of
pretrained visual representations for
control~\cite{nair2023r3m,majumdar2023vc1}.
For each camera view $c$, we average-pool its feature map
to a fixed spatial grid and flatten the result:
$h_t^{(c)}
=\operatorname{vec}(
\operatorname{Pool}(\phi(I_t^{(c)})))$.
This retains coarse spatial information while reducing
feature dimensionality.

To obtain compact inputs, we fit a separate
principal component analysis (PCA)
transformation~\cite{jolliffe2016pca} $P_{\tau}^{(c)}$ for each
camera using observations pooled across all $K$ supplied
demonstrations. Each transformation includes centering
by the corresponding demonstration feature mean.
The resulting features are
$z_t^{(c)}=P_{\tau}^{(c)}(h_t^{(c)})$.
We concatenate these projected features with robot
proprioception to form the numerical input:
\begin{equation}
x_t =
\left[
z_t^{(1)};\ldots;z_t^{(C)};s_t
\right]
\in\mathbb{R}^{d_x},
\label{eq:numerical_observation}
\end{equation}
where $d_x$ is the resulting input dimension.
The same projections encode demonstration and query images;
they use no observations from evaluation rollouts and
remain fixed during execution. When the demonstration set
changes, we refit the projections and rebuild the context
while keeping the encoder frozen.

\paragraph{Action-sequence targets.}
\label{sec:action_representation}

We represent future action sequences as numerical targets,
following action-chunk prediction in imitation learning
\cite{zhao2023act,chi2023diffusionpolicy}.
For a prediction horizon $H$, each observation is
associated with
$A_t=[a_t,\ldots,a_{t+H-1}]$.
Chunks stay within each demonstration, with its final
action repeated when fewer than $H$ actions remain.

To reduce the number of scalar prediction targets, we
preserve the immediate continuous action and compress
subsequent motion using PCA. The immediate action is therefore unaffected by PCA
reconstruction.
Let
$r_t=\operatorname{vec}(u_{t+1},\ldots,u_{t+H-1})$
denote the future continuous-motion sequence, and
let $\rho_{\tau}$ be a PCA transformation fitted on these
sequences from all supplied demonstrations.
Writing its mean as $\mu_{\tau}$ and its $d_p$ retained
principal directions as the columns of $B_{\tau}$,
the future-motion coefficients are
\begin{equation}
c_t=\rho_{\tau}(r_t)
=B_{\tau}^{\top}(r_t-\mu_{\tau})
\in\mathbb{R}^{d_p}.
\label{eq:action_encoding}
\end{equation}
The numerical target is
\begin{equation}
y_t =
\left[
u_t;c_t;
g_t;\ldots;g_{t+H-1}
\right].
\label{eq:action_target}
\end{equation}
The immediate motion components and future-motion PCA
coefficients are regression targets. Gripper commands
are classification targets over the discrete values
observed in the supplied demonstrations.
Thus, for $d_u$ continuous motion dimensions, each
chunk has $d_y=d_u+d_p+H$ scalar targets rather than
$H(d_u+1)$ raw action components: 26 instead of 56
in our default LIBERO configuration.
Because the basis spans both future time steps and motion
dimensions, each coefficient describes a motion pattern
coordinated across them.
For $H=1$, the future-motion component is omitted.

\subsection{Frozen Prediction and Closed-Loop Execution}
\label{sec:inference}

\paragraph{Preparing demonstration context.}
Before execution, we fit the visual and action
representations on the supplied demonstrations and
encode each demonstration time step as one numerical
input--output pair. We collect all such pairs into
the context:
\begin{equation}
\mathcal{C}_{\tau}
=
\{(x_i,y_i)\}_{i=1}^{N},
\label{eq:tabular_context}
\end{equation}
where $N=\sum_{k=1}^{K}T_k$ and $i$ indexes time steps
across demonstrations. Rows from all demonstrations are pooled
without temporal alignment: each input describes a single
observation, and temporal structure enters only through
its action-chunk target.

\paragraph{Predicting and decoding actions.}
We instantiate the predictor with frozen
TabICLv2~\cite{tabiclv2}. Given the current
input $x_t$, we query it separately for each
scalar target using the corresponding context labels:
\begin{equation}
\hat{y}_t^{(j)}
=
f_{\theta}^{(j)}
\left(
x_t
\mid
\{(x_i,y_i^{(j)})\}_{i=1}^{N}
\right).
\label{eq:tfm_prediction}
\end{equation}
Here, $f_{\theta}^{(j)}$ denotes the frozen regression
or classification predictor applied to target $j$.
Continuous motion and PCA coefficients use the regression
checkpoint, while gripper targets use the classification
checkpoint. Parameters are shared within each target type.
We reconstruct future motion from the predicted
coefficients $\hat{c}_t$:
\begin{equation}
\hat{r}_t=\mu_{\tau}+B_{\tau}\hat{c}_t.
\label{eq:action_decoding}
\end{equation}
Reshaping $\hat{r}_t$ yields the next $H-1$ continuous
motion commands. Although each coefficient is predicted
independently, decoding through the shared basis keeps the
reconstructed motion within the principal motion patterns
of the demonstrations.
We map predicted gripper classes back to their command values.
Combining these with the predicted immediate motion
yields $\hat{A}_t=[\hat{a}_t,\ldots,\hat{a}_{t+H-1}]$.

\paragraph{Closed-loop execution.}
The robot executes the
first $E\leq H$ predicted actions, obtains a new
observation, and repeats the prediction process.
The demonstration context and PCA transformations
remain fixed during execution, while each new
observation forms a new query.
For $E<H$, the remaining actions are discarded
when the next observation arrives.

Deployment thus requires context preparation and PCA
fitting, but no gradient-based policy optimization; the
TFM and visual encoder remain frozen throughout.

\begin{table*}[t]
\centering
\caption{
\textbf{Robot imitation on unseen manipulation tasks.}
Success rates (\%) on held-out LIBERO and Meta-World tasks.
Methods use one target demonstration for in-context adaptation
(including retrieval) or fine-tuning (+1-shot FT); pretrained
robot policies without target fine-tuning use none.
LIBERO Avg. is the mean across the Object, Spatial, and Goal
subsets. Best results are bold.
}
\label{tab:main_results}

\setlength{\tabcolsep}{5pt}
\renewcommand{\arraystretch}{1.15}
\small

\begin{tabular}{lcccccccc}
\toprule

\multirow[c]{2}{*}[-0.4ex]{\textbf{Method}}
& \multirow[c]{2}{*}[-0.4ex]{
    \shortstack[c]{\textbf{No Source-Task}\\\textbf{Training}}
}
& \multirow[c]{2}{*}[-0.4ex]{
    \shortstack[c]{\textbf{In-Context}\\\textbf{Adaptation}}
}
& \multirow[c]{2}{*}[-0.4ex]{
    \shortstack[c]{\textbf{No Target}\\\textbf{FT}}
}
& \multicolumn{4}{c}{\textbf{LIBERO}}
& \multirow[c]{2}{*}[-0.4ex]{\textbf{Meta-World}} \\

\cmidrule(lr){5-8}

& & & & Object & Spatial & Goal & Avg. & \\

\midrule

\multicolumn{9}{l}{
    \textit{Robot-Trained In-Context Learning}
} \\

ICRT~\cite{fu2024icrt}
& $\times$ & \checkmark & \checkmark
& 64.2 & 4.9 & 0.9 & 23.3 & 4.3 \\

ICLR~\cite{nguyen2026iclr}
& $\times$ & \checkmark & \checkmark
& 69.3 & 2.2 & 0.0 & 23.9 & 4.5 \\

ContextFlow~\cite{contextflow}
& $\times$ & \checkmark & \checkmark
& \textbf{73.6} & 4.7 & 3.3 & 27.2 & 19.6 \\

\midrule

\multicolumn{9}{l}{
    \textit{Retrieval-Based Imitation}
} \\

VINN~\cite{pari2022vinn}
& $\times$ & \checkmark & \checkmark
& 53.3 & 15.6 & 8.2 & 25.7 & 39.1 \\

R\&P~\cite{sridhar2025regent}
& \checkmark & \checkmark & \checkmark
& 49.3 & 28.0 & 48.0 & 41.8 & 38.1 \\

\midrule

\multicolumn{9}{l}{
    \textit{Pretrained Robot Policies}
} \\

$\pi_{0.5}$~\cite{black2025pi05}
& $\times$ & $\times$ & \checkmark
& 9.3 & 52.0 & 18.7 & 26.7 & 2.4 \\

$\pi_{0.5}$~\cite{black2025pi05} (+1-shot FT)
& $\times$ & $\times$ & $\times$
& 23.8 & \textbf{56.9} & 44.4 & 41.7 & 25.2 \\

GR00T N1.5~\cite{nvidia2025groot15}
& $\times$ & $\times$ & \checkmark
& 2.0 & 6.7 & 2.7 & 3.8 & 0.8 \\

GR00T N1.5~\cite{nvidia2025groot15} (+1-shot FT)
& $\times$ & $\times$ & $\times$
& 34.4 & 42.0 & 45.6 & 40.7 & 43.5 \\

\midrule

\rowcolor{gray!10}
\textbf{\ours{} (TabICLv2)}
& \checkmark & \checkmark & \checkmark
& 70.2 & 42.2 & \textbf{75.6}
& \textbf{62.7} & \textbf{49.3} \\

\bottomrule
\end{tabular}
\end{table*}

\section{Experiments}
\label{sec:experiments}

\subsection{Experimental Setup}
\label{sec:experimental_setup}

\paragraph{Benchmarks and held-out tasks.}
We evaluate visual one-shot imitation on held-out tasks from
LIBERO~\cite{liu2023libero} and Meta-World~\cite{yu2020metaworld},
following the evaluation setting of ICLR~\cite{nguyen2026iclr}.
Source tasks are used to train robot-policy baselines
and the VINN visual representation; target tasks are excluded
from this training, and their demonstrations are provided
only at evaluation time.
For LIBERO-Object, we follow ICLR's seven-source/three-held-out
task split, and we apply the same structure to
LIBERO-Spatial and LIBERO-Goal. All reported suite
results refer to these held-out subsets.
For Meta-World, we use 45 source and five held-out task
classes for visual imitation. Observations provide RGB images and robot
proprioception; no method receives privileged object or
goal states. Our predictor uses no source-task demonstrations
on either benchmark.

\paragraph{Evaluation protocol.}
Each one-shot evaluation condition uses exactly one
target-task demonstration, either as in-context input
or for fine-tuning, depending on the method.
Following ICLR, we evaluate three independent one-shot
demonstration conditions per held-out LIBERO task, each
over the same 50 initial states, giving 450 rollouts per subset.
LIBERO Avg.\ is the unweighted mean of the Object,
Spatial, and Goal subset success rates.
For Meta-World, we evaluate five scene configurations per
held-out task class and ten perturbed robot initial positions
within each scene, keeping its layout and goal fixed.
Across five held-out classes and three independent one-shot
demonstrations per scene, this yields 750 rollouts.
Corresponding one-shot comparisons use the same target
tasks, demonstration identities, and query initializations.

\paragraph{Baselines.}
We compare with three groups of methods; details are in
Appendix~\ref{sec:supp_baselines}.
\emph{Robot-trained ICL models}, ICRT~\cite{fu2024icrt},
ICLR~\cite{nguyen2026iclr}, and ContextFlow~\cite{contextflow},
are trained on source tasks and use a target-task
demonstration as context without test-time parameter updates.
\emph{Retrieval-based methods} use the demonstration as an
action memory. VINN~\cite{pari2022vinn} uses a BYOL visual
encoder trained on source tasks. R\&P~\cite{sridhar2025regent},
the retrieval component of REGENT rather than the full system,
uses the ImageNet-pretrained ResNet-18
encoder from the REGENT implementation.
Both retrieve single-step actions from RGB images with
frozen encoders: VINN uses ten neighbors selected by source
validation; R\&P uses one.
\emph{Pretrained robot policies}, $\pi_{0.5}$~\cite{black2025pi05}
and GR00T N1.5~\cite{nvidia2025groot15}, are fine-tuned from
their foundation checkpoints on source tasks, then evaluated
directly without target demonstrations or after fine-tuning
on one target-task demonstration (+1-shot FT).

\paragraph{Implementation details.}
Our default configuration uses frozen DINOv2-S/14
features~\cite{oquab2024dinov2} and
TabICLv2~\cite{tabiclv2}, with up to 64 visual PCA
components per camera and 12 future-motion PCA components.
Unless otherwise stated, experiments using our tabular
interface use prediction and execution horizons $H=E=8$
on both benchmarks. Projections are fitted only on the supplied
demonstration and remain fixed during rollout.
Predicting one action chunk takes a median of 881\,ms
(mean 821\,ms) on an RTX 4090 at batch size one, with
amortized throughput of 9.6\,Hz, excluding context
preparation and simulation.

\subsection{One-Shot Imitation Results}
\label{sec:main_results}

We conduct two complementary comparisons.
Table~\ref{tab:main_results} evaluates complete robot
imitation methods, each with its own representation
and action-generation mechanism.
Table~\ref{tab:numerical_learners} then fixes the
visual features, proprioception, action targets,
demonstration context, and execution protocol, and
varies only the predictor.

\paragraph{Comparison with complete methods.}
From one demonstration, \ours{} with frozen TabICLv2 achieves
the highest average success on both benchmarks: 62.7\% on
LIBERO and 49.3\% on Meta-World
(Table~\ref{tab:main_results}).

Among the robot-trained ICL models, ContextFlow is the
strongest, with 27.2\% on LIBERO and 19.6\% on Meta-World.
These models are competitive on Object, where ContextFlow
reaches 73.6\% against our 70.2\%, but none exceeds 4.9\%
on Spatial or Goal. Our gains over them are therefore
concentrated on these two subsets.
With demonstrations of their source tasks from the same
initial states, these models achieve 85.7--90.0\% success
on LIBERO-Goal (Appendix~\ref{sec:supp_source_controls}),
so their low held-out Goal success reflects difficulty
following demonstrations of unseen goals, which
Sec.~\ref{sec:context_behavior} examines.

Retrieval provides a stronger reference for using
demonstrations directly at deployment. R\&P reaches 41.8\%
on LIBERO and 38.1\% on Meta-World, above all three
robot-trained ICL models on both averages; VINN reaches
25.7\% and 39.1\%. \Ours{} exceeds both retrieval methods
on every reported subset, and exceeds the stronger retrieval
result on each benchmark average by 20.9 percentage points
on LIBERO and 10.2 on Meta-World.

The pretrained robot policies reach at most 41.7\% on LIBERO
and 43.5\% on Meta-World after one-shot fine-tuning, and
less without it. One-shot fine-tuned $\pi_{0.5}$ attains
the highest Spatial success (56.9\%), so our advantage does
not hold on every subset. On average, however, a frozen,
generically pretrained predictor exceeds the evaluated
methods, including those that train on source tasks or
fine-tune on the target demonstration.

\paragraph{Predictors under a shared interface.}
We next ask whether the predictor matters once the examples
are fixed. Under the default settings, we replace TabICLv2
with kNN, linear models, or CatBoost while keeping the
visual features, proprioception, context, action targets,
decoding, and execution horizons unchanged.
Each conventional predictor is fitted only on the supplied
target demonstration, with hyperparameters selected on
source-task validation trajectories
(Appendix~\ref{sec:supp_predictor_selection}).

\begin{table}[htbp]
\centering
\caption{
\textbf{Predictor comparison under a shared interface.}
One-shot success (\%) with fixed observation
features, action targets, demonstrations, and query
initializations. Conventional predictors are tuned on
source-task validation trajectories.
}
\label{tab:numerical_learners}

\setlength{\tabcolsep}{3pt}
\renewcommand{\arraystretch}{1.12}
\footnotesize
\begin{tabular*}{\columnwidth}{
@{\extracolsep{\fill}}lccccc@{}}
\toprule
\multirow[c]{2}{*}[-0.4ex]{\textbf{Predictor}}
& \multicolumn{4}{c}{\textbf{LIBERO}}
& \multirow[c]{2}{*}[-0.4ex]{\textbf{Meta-World}} \\
\cmidrule(lr){2-5}
& Object & Spatial & Goal & Avg. & \\
\midrule
kNN (tuned)
& 65.8 & 20.4 & 34.0 & 40.1 & 14.1 \\
Linear (tuned)
& 2.9 & 21.6 & 26.0 & 16.8 & 44.1 \\
CatBoost (tuned)
& 0.2 & 0.0 & 6.4 & 2.2 & 16.1 \\
\midrule
\textbf{TabICLv2}
& \textbf{70.2} & \textbf{42.2} & \textbf{75.6}
& \textbf{62.7} & \textbf{49.3} \\
\bottomrule
\end{tabular*}
\end{table}

TabICLv2 achieves the highest average success on both
benchmarks (Table~\ref{tab:numerical_learners}).
No conventional predictor is strong on both: kNN reaches
40.1\% on LIBERO but 14.1\% on Meta-World, whereas Linear
reaches 44.1\% on Meta-World but 16.8\% on LIBERO.
The closest results are kNN on Object (65.8\% versus 70.2\%)
and Linear on Meta-World (44.1\% versus 49.3\%).
Under this shared interface, the pretrained TFM is thus the
only tested predictor that performs well on both benchmarks.

\paragraph{Effect of the action horizon.}
\begin{table}[htbp]
\centering
\caption{
\textbf{Effect of the action horizon.}
One-shot success (\%) on held-out LIBERO tasks when
prediction and execution horizons vary together ($H=E$);
all other inputs are fixed. kNN uses one neighbor at both
horizons, matching its source-selected LIBERO setting in
Table~\ref{tab:numerical_learners}. Bold marks the better
predictor at each horizon.
}
\label{tab:action_horizon}

\setlength{\tabcolsep}{3pt}
\renewcommand{\arraystretch}{1.12}
\footnotesize
\begin{tabular*}{\columnwidth}{@{\extracolsep{\fill}}lccccc@{}}
\toprule
\textbf{Predictor} & $H=E$ & Object & Spatial & Goal & Avg. \\
\midrule
kNN & 1 & \textbf{57.1} & 22.2 & 24.4 & 34.6 \\
kNN & 8 & 65.8 & 20.4 & 34.0 & 40.1 \\
\midrule
TabICLv2 & 1 & 39.3 & \textbf{35.1} & \textbf{55.6} & \textbf{43.3} \\
TabICLv2 & 8 & \textbf{70.2} & \textbf{42.2} & \textbf{75.6} & \textbf{62.7} \\
\bottomrule
\end{tabular*}
\end{table}

Table~\ref{tab:action_horizon} repeats the comparison
under single-step control ($H=E=1$) for TabICLv2 and kNN,
using one neighbor for kNN in both LIBERO horizon settings.
TabICLv2 retains the higher average success
(43.3\% versus 34.6\%), although kNN performs better on
Object (57.1\% versus 39.3\%). Moving to eight-step
prediction and execution improves average success by 19.4
percentage points for TabICLv2 and 5.5 for kNN, so the gap
between predictors widens from 8.7 to 22.6 points. Because prediction and execution
horizons change together, these gains reflect their combined
effect; Appendix~\ref{sec:supp_action_horizons} reports
additional horizon settings.
These results suggest that temporal action chunking is
particularly effective when paired with TabICLv2, amplifying
its average advantage over nearest-neighbor retrieval.

\subsection{Generality Across Model Configurations}
\label{sec:tfm_generality}

\begin{table}[htbp]
\centering
\caption{\textbf{Generality across tabular foundation models.}
One-shot success (\%) of \ours{} with frozen DINOv2-S/14 features
and up to 64 visual PCA components per camera.}
\label{tab:tfm_generality}

\setlength{\tabcolsep}{3pt}
\renewcommand{\arraystretch}{1.12}
\footnotesize

\begin{tabular*}{\columnwidth}{
@{\extracolsep{\fill}}lccccc@{}}
\toprule
\multirow[c]{2}{*}[-0.4ex]{\textbf{Model}}
& \multicolumn{4}{c}{\textbf{LIBERO}}
& \multirow[c]{2}{*}[-0.4ex]{\textbf{Meta-World}} \\
\cmidrule(lr){2-5}
& Object & Spatial & Goal & Avg. & \\
\midrule

TabPFN-3.5-Fast~\cite{jaeger2026tabpfn35}
& 55.8 & 38.9 & 42.4 & 45.7 & 38.0 \\

Kumo Tabular-S~\cite{nvidia2026kumo}
& 59.6 & 47.3 & 68.4 & 58.4 & 40.8 \\

Kumo Tabular-M~\cite{nvidia2026kumo}
& 65.3 & 46.7 & 72.4 & 61.5 & 41.1 \\

Causilo~\cite{cho2026causilo}
& 62.4 & \textbf{48.7}
& \textbf{78.2} & \textbf{63.1} & 40.3 \\

\midrule
\rowcolor{gray!10}
\textbf{TabICLv2} (default)
& \textbf{70.2} & 42.2
& 75.6 & 62.7 & \textbf{49.3} \\

\bottomrule
\end{tabular*}
\end{table}

We next examine whether effective control depends on a
particular TFM or visual encoder, varying one component
at a time while keeping the other fixed.

\paragraph{Across tabular foundation models.}
We replace TabICLv2 with
TabPFN-3.5-Fast~\cite{jaeger2026tabpfn35},
Kumo Tabular Small and Medium~\cite{nvidia2026kumo},
or Causilo~\cite{cho2026causilo}, keeping the visual
encoder, inputs and targets, demonstrations, and evaluation
protocol fixed. All predictors remain frozen.

Across the five model configurations in
Table~\ref{tab:tfm_generality}, average success ranges
from 45.7\% to 63.1\% on LIBERO and from 38.0\% to
49.3\% on Meta-World. Every configuration exceeds
both evaluated retrieval methods in average LIBERO
success, and all but TabPFN-3.5-Fast do so on Meta-World.
Causilo has the highest LIBERO average (63.1\%), close
to TabICLv2 (62.7\%), while TabICLv2 achieves the
highest Object and Meta-World success.
Effective in-context imitation thus extends across multiple
pretrained TFMs rather than depending on a single model.

\paragraph{Across visual encoders and projections.}
\begin{table}[htbp]
\centering
\caption{\textbf{Effect of the visual encoder.}
One-shot success (\%) of \ours{} with frozen TabICLv2
and up to 64 visual PCA components per camera.
Only the frozen visual encoder changes.}
\label{tab:visual_encoders}

\setlength{\tabcolsep}{3pt}
\renewcommand{\arraystretch}{1.12}
\footnotesize

\begin{tabular*}{\columnwidth}{
@{\extracolsep{\fill}}lccccc@{}}
\toprule
\multirow[c]{2}{*}[-0.4ex]{\textbf{Visual encoder}}
& \multicolumn{4}{c}{\textbf{LIBERO}}
& \multirow[c]{2}{*}[-0.4ex]{\textbf{Meta-World}} \\
\cmidrule(lr){2-5}
& Object & Spatial & Goal & Avg. & \\
\midrule
DINOv2-S/14~\cite{oquab2024dinov2} (default)
& 70.2 & 42.2 & \textbf{75.6} & 62.7 & 49.3 \\
SigLIP2-B/16~\cite{tschannen2025siglip2}
& 64.7 & 44.9 & 67.3 & 59.0 & 50.4 \\
MAE-B/16~\cite{he2022mae}
& 54.7 & 40.0 & 62.9 & 52.5 & 45.6 \\
DINOv3-S/16~\cite{simeoni2025dinov3}
& 66.4 & \textbf{50.2} & 63.8 & 60.1 & 50.3 \\
DINOv3-B/16~\cite{simeoni2025dinov3}
& \textbf{78.4} & 46.7 & 68.4 & \textbf{64.5} & \textbf{52.9} \\

\bottomrule
\end{tabular*}
\end{table}

Table~\ref{tab:visual_encoders} compares five
frozen visual encoders under the same prediction
formulation. Average LIBERO success ranges from
52.5\% with MAE-B~\cite{he2022mae} to 64.5\% with
DINOv3-B~\cite{simeoni2025dinov3}; Meta-World success
ranges from 45.6\% to 52.9\%.
The best encoder varies across subsets, but effective
control is not specific to a single visual backbone.
The projection matters more than its size: with
DINOv2-S/14 fixed, varying visual PCA dimensions
from 16 to 64 yields similar LIBERO averages (62.7--63.2\%),
while replacing PCA-64 with an equal-dimensional random
projection reduces success to 53.6\%
(Appendix~\ref{sec:supp_visual_projection}).

\subsection{Demonstration Conditioning and Scaling}
\label{sec:context_behavior}

Finally, we test two behaviors expected of an in-context
learner: changing the demonstration should change the
pursued goal, and more demonstrations should improve success.

\paragraph{Changing the demonstrated goal.}
\label{sec:goal_conditioning}

We conduct a controlled goal-switching experiment
on three held-out LIBERO-Goal tasks
in the same scene: opening a drawer, placing
a bowl on a cabinet, and placing a bowl on a plate
(Fig.~\ref{fig:goal_switch}).
Initial observations are identical across task conditions,
so only the supplied demonstration distinguishes the tasks
for \ours{}, ICRT, ICLR, VINN, and R\&P;
ContextFlow additionally receives the corresponding
language instruction.
We evaluate three separate one-shot demonstration
conditions per task, each from the same 20 initial
states, giving 180 rollouts per method. Each rollout runs
for 300 steps without early termination on success, while
we monitor the goal predicates of all ten tasks
in the suite.

\begin{table}[htbp]
\centering
\caption{
\textbf{Goal-conditioned behavior under identical
initial observations.}
\emph{Requested}: success rate (\%) on three held-out
Goal tasks. \emph{Source goal}: number of the same 180
rollouts that complete a source-task goal without
completing the requested or another held-out goal.
}
\label{tab:goal_switch}

\setlength{\tabcolsep}{2pt}
\renewcommand{\arraystretch}{1.12}
\small

\begin{tabular*}{\columnwidth}{
@{\extracolsep{\fill}}lcc@{}}
\toprule
\textbf{Method} & Requested (\%) $\uparrow$ & Source goal (count) $\downarrow$ \\
\midrule
ICRT~\cite{fu2024icrt}         & 0.6 & 27 \\
ICLR~\cite{nguyen2026iclr}     & 0.0 & 5  \\
ContextFlow~\cite{contextflow} & 4.4 & 87 \\
\midrule
VINN~\cite{pari2022vinn} & 7.2 & \textbf{0} \\
R\&P~\cite{sridhar2025regent} & 48.3 & \textbf{0} \\
\midrule
\textbf{\ours{} (TabICLv2)}
& \textbf{75.6} & \textbf{0} \\
\bottomrule
\end{tabular*}
\end{table}

\Ours{} achieves 75.6\% requested-goal success,
compared with 48.3\% for R\&P, 7.2\% for VINN, and
0.0--4.4\% for the robot-trained ICL methods
(Table~\ref{tab:goal_switch}). R\&P's success shows
that goal switching is also possible through retrieval;
the TFM attains higher success in this controlled setting.
With source-task demonstrations and the same initial states,
the robot-trained ICL models achieve 85.7--90.0\% success
on source tasks (Appendix~\ref{sec:supp_source_controls}).
The failures also differ in kind. The robot-trained ICL
methods complete a source-task goal instead in 5 to 87
of their 180 rollouts, whereas neither \ours{} nor either
retrieval method completes a different held-out or source
goal on its unsuccessful rollouts.
With identical initial observations and no language input,
these results provide evidence that the behavior of \ours{}
is conditioned on the supplied demonstration.

\paragraph{Learning from more demonstrations.}
\label{sec:context_scaling}

We evaluate $K\in\{1,3,5\}$ demonstrations across the
Object, Spatial, and Goal held-out subsets, using three nested
demonstration groups and fixed query initial states.
Figure~\ref{fig:teaser} illustrates the LIBERO-Object results,
where \ours{} improves from 70.2\% success with one
demonstration to 94.2\% with five.
\Ours{} pools the demonstrations' observation--action examples and
refits the PCA projections, keeping pretrained weights
frozen. R\&P enlarges its retrieval memory with the same
demonstrations and a fixed neighbor count; $K$ counts
demonstrations, not neighbors. ContextFlow is evaluated
only at $K=1$ because its released implementation accepts
a single demonstration.

Table~\ref{tab:context_scaling_main} shows that the LIBERO
average success of \ours{} rises from 62.7\% with one
demonstration to 81.3\% with five, exceeding the compared
baselines at each $K$. Both retrieval methods also improve,
while ICRT and ICLR show no consistent gains; for these
two models, contexts with $K\geq3$ often exceed the context
length used in training.
The gains also extend to Meta-World, where \ours{} improves
from 49.3\% to 84.5\% and outperforms both retrieval methods
at each $K$.

\begin{table}[htbp]
\centering
\caption{\textbf{Learning from more demonstrations.}
Success rates (\%) use 450 rollouts per LIBERO subset
and 750 on Meta-World at each $K$. LIBERO averages
Object, Spatial, and Goal; Meta-World scaling was not
evaluated for the robot-trained ICL methods.}
\label{tab:context_scaling_main}
\setlength{\tabcolsep}{4pt}
\small
\begin{tabular*}{\columnwidth}{@{\extracolsep{\fill}}lccc@{}}
\toprule
\textbf{Method} & $K=1$ & $K=3$ & $K=5$ \\
\midrule
\multicolumn{4}{c}{\textbf{LIBERO Avg.}} \\
\midrule
ICRT & 23.3 & 25.0 & 19.3 \\
ICLR & 23.9 & 27.0 & 23.1 \\
\midrule
VINN & 25.7 & 46.2 & 50.2 \\
R\&P & 41.8 & 53.6 & 56.8 \\
\midrule
\textbf{\ours{} (TabICLv2)} & \textbf{62.7} & \textbf{75.7} & \textbf{81.3} \\
\midrule
\multicolumn{4}{c}{\textbf{Meta-World}} \\
\midrule
VINN & 39.1 & 63.1 & 72.4 \\
R\&P & 38.1 & 57.2 & 65.5 \\
\textbf{\ours{} (TabICLv2)} & \textbf{49.3} & \textbf{79.6} & \textbf{84.5} \\
\bottomrule
\end{tabular*}
\end{table}

\section{Conclusion and Discussion}
\label{sec:conclusion}

We introduced \ours{}, which casts visual imitation as
numerical prediction for frozen TFMs, enabling closed-loop
control without robot-specific policy training or target-task
fine-tuning. On the evaluated held-out LIBERO and Meta-World
tasks, \ours{} with TabICLv2 achieves the highest one-shot
average success among the tested methods.
Demonstration-conditioned goal switching, gains from
additional demonstrations, and effective control across
multiple TFMs and visual encoders support generic numerical
pretraining as a viable foundation for in-context robot learning.

\paragraph{Limitations and future work.}
Our evaluation is limited to simulation, and inference
latency constrains deployment.
Inputs contain only the current observation; incorporating
observation history could support tasks that require memory.
An exploratory multi-output numerical model reduces
action-chunk prediction latency (Appendix~\ref{sec:efficiency}).
Future work should assess whether faster prediction preserves
control performance and enables effective deployment on
physical robots.

{
    \small
    \bibliographystyle{ieeenat_fullname}
    \bibliography{main}
}

\clearpage
\appendix
\twocolumn[
  \centering
  \Large
  \textbf{Appendix}\\
  \vspace{1.0em}
]
\setcounter{section}{0}
\renewcommand{\thesection}{\Alph{section}}
\renewcommand{\theHsection}{supp.\Alph{section}}
\setcounter{table}{0}
\setcounter{figure}{0}
\renewcommand{\thetable}{S\arabic{table}}
\renewcommand{\thefigure}{S\arabic{figure}}
\renewcommand{\theHtable}{supp.\arabic{table}}
\renewcommand{\theHfigure}{supp.\arabic{figure}}

\section{Detailed Experimental Setup}
\label{sec:supp_setup}

\paragraph{Overview.}
We study in-context imitation on held-out tasks of
LIBERO~\cite{liu2023libero} and Meta-World~\cite{yu2020metaworld},
following the evaluation setting of ICLR~\cite{nguyen2026iclr}.
ICLR instantiates this setting on LIBERO; on Meta-World,
we construct a protocol with the same structure from the
ML45 class split, described below.
Source tasks are used to train the robot-policy baselines
and the VINN visual representation; each such method trains
one model per LIBERO suite on that suite's source tasks and
one model on the Meta-World source classes. Target-task demonstrations are provided only at
evaluation time, and our numerical predictor uses no
source-task demonstrations on either benchmark.

\paragraph{Observations and actions.}
On LIBERO, observations consist of agent-view and
wrist-camera images rendered at $256\times256$ and
8-dimensional proprioception: end-effector position,
axis--angle orientation, and the two gripper finger
positions. Actions contain end-effector position and
axis--angle orientation deltas and a gripper command.
On Meta-World, observations consist of images from a
single fixed third-person camera rendered at
$480\times480$ and 4-dimensional proprioception:
end-effector position and gripper opening. Actions contain
end-effector position deltas and a gripper command.
The retrieval methods use only the images, while
ContextFlow, $\pi_{0.5}$, and GR00T N1.5 additionally
receive the task's language instruction. No method
receives privileged object or goal states.

\begin{table*}[t]
\centering
\caption{\textbf{LIBERO task split and one-shot demonstrations.}
Task IDs follow the default task order of each LIBERO suite; the seven
tasks not held out are source tasks. Demonstration IDs index the public
LIBERO dataset of the task; each ID defines one independent one-shot condition.}
\label{tab:task_split}
\small
\begin{tabular}{llcll}
\toprule
Suite & Source task IDs & ID & Held-out task & Demonstration IDs \\
\midrule
Object & 1, 3, 4, 6, 7, 8, 9
 & 0 & pick up the alphabet soup and place it in the basket & 49, 10, 12 \\
 & & 2 & pick up the salad dressing and place it in the basket & 41, 1, 34 \\
 & & 5 & pick up the tomato sauce and place it in the basket & 38, 10, 13 \\
\midrule
Spatial & 0, 1, 2, 4, 6, 7, 9
 & 3 & pick up the black bowl on the cookie box and place it on the plate & 41, 0, 22 \\
 & & 5 & pick up the black bowl on the ramekin and place it on the plate & 40, 22, 49 \\
 & & 8 & pick up the black bowl next to the plate and place it on the plate & 2, 25, 36 \\
\midrule
Goal & 1, 2, 3, 5, 6, 7, 9
 & 0 & open the middle drawer of the cabinet & 25, 23, 37 \\
 & & 4 & put the bowl on top of the cabinet & 28, 19, 49 \\
 & & 8 & put the bowl on the plate & 2, 26, 38 \\
\bottomrule
\end{tabular}
\end{table*}

\paragraph{LIBERO.}
Each suite contains ten tasks, of which three are held out
and the remaining seven are source tasks
(Table~\ref{tab:task_split}); all reported suite results
refer to these held-out subsets.
For Object, we follow ICLR's seven-source/three-held-out split.
For Spatial, tasks 3 and 8 are the two Spatial tasks held
out in the ContextFlow release, and we add task 5 to obtain
three held-out tasks. For Goal, we hold out tasks 0, 4, and 8.
Source-task training uses the demonstrations of the public
LIBERO dataset (50 per task), replayed in simulation
following ICLR's preprocessing; demonstrations that do not
replay successfully are discarded.

Following ICLR, each held-out task has three independent
one-shot conditions. Their demonstrations are sampled
without replacement, using a fixed random seed, from the
demonstrations of that task that replay successfully
(Table~\ref{tab:task_split}).
Each condition is evaluated on the same 50 official initial
states, giving $3\times3\times50=450$ rollouts per subset,
with at most 300 environment action steps per rollout.

\paragraph{Meta-World.}
We adopt the class split of the Meta-World ML45 benchmark
for visual imitation: its 45 training classes are the source
classes, and its five test classes (bin-picking, box-close,
hand-insert, door-lock, and door-unlock) are held out.
All demonstrations are generated with the benchmark's
scripted expert policies, keeping only successful rollouts.
For source-task training, each source class has twelve
scenes with different object and goal placements and five
perturbed robot initial positions per scene, giving 60 demonstrations
per class. Ten scenes per class are used for training and
the remaining two for source validation, yielding 2,250
training and 450 validation demonstrations across the
45 source classes.

Each held-out class has five scenes with different object
and goal positions, sampled with fixed seeds as in
Meta-World's task variants. As in Meta-World's definition
of a task instance, positions are fixed within a scene,
and each scene has its own three one-shot demonstrations;
unlike Meta-World's fixed robot initialization, the
robot's initial end-effector position is perturbed.
Candidate initial positions are enumerated in a fixed order,
and the first three on which the scripted expert succeeds
provide the three independent one-shot demonstrations.
Ten further initial positions, disjoint from the
demonstration candidates, serve as queries.
Each demonstration is evaluated on the ten queries of its
scene, giving $5\times5\times3\times10=750$ rollouts,
with at most 500 environment action steps per rollout.

\paragraph{Shared protocol.}
Each one-shot evaluation condition uses exactly one
target-task demonstration, either as in-context input
or for fine-tuning, depending on the method.
A rollout is successful if the benchmark's success
condition is met within the step limit, and it ends as
soon as the condition is met.
Corresponding one-shot comparisons use the same target
tasks, demonstration identities, and query initializations.
We report aggregate success rates for each LIBERO subset
and for the evaluated Meta-World tasks; LIBERO Avg.\ is
the unweighted mean of the Object, Spatial, and Goal
subset success rates.

\section{Implementation of \ours{}}
\label{sec:supp_implementation}

\paragraph{Visual features.}
Meta-World frames are first resized to $256\times256$ to
match LIBERO, and all frames are then resized to
$224\times224$ with bilinear, antialiased interpolation,
followed by ImageNet normalization.
Frozen DINOv2-S/14~\cite{oquab2024dinov2}, run under
bfloat16 autocast, produces a $16\times16$ map of
384-dimensional patch tokens after its final layer
normalization; the class token is discarded. We
average-pool the map over a $4\times4$ grid of
$4\times4$-token cells and flatten the result, giving
$4\times4\times384=6{,}144$ features per camera.

\paragraph{Default configuration.}
The default configuration applies TabICLv2~\cite{tabiclv2}
to these features. For each camera, PCA fitted on the
supplied demonstration retains up to 64 components.
A demonstration with $n$ frames gives centered features
of rank at most $n-1$, so each camera retains
$\min(64,n-1)$ components. All one-shot LIBERO
demonstrations contain 91 to 177 frames and retain all
64 components; the Meta-World demonstrations contain 39
to 144 frames, and 27 of the 75 retain fewer than 64.
LIBERO inputs thus have 136 features (two views and
8-dimensional proprioception), and Meta-World inputs at
most 68 (one view and 4-dimensional proprioception).

Both benchmarks use a prediction horizon $H=8$ and
execution horizon $E=8$. Future motion is projected onto
12 PCA components fitted on the same demonstration.
LIBERO's targets comprise six immediate motion components,
12 future-motion coefficients, and eight gripper targets
(26 in total); Meta-World's comprise three, 12, and eight
(23). Both projections remain fixed during rollout.

\paragraph{TabICLv2 prediction details.}
We use the authors' frozen TabICLv2 checkpoints, with the
regression model shared by all motion targets and the
classification model shared by all gripper targets;
a gripper target with a single class in the demonstration
returns that class directly.
Context preparation encodes the demonstration rows, with
each target's own labels, into a per-target cache, using
four ensemble estimators and TabICLv2's default
preprocessing; the cache is reused for
all queries during rollout.
Regression targets take the predictive mean, computed as
the mean of the predicted quantiles, and gripper targets
take the most probable class. After inverse PCA and
gripper decoding, actions are clipped to the action range
$[-1,1]$.

\section{Baselines}
\label{sec:supp_baselines}

We compare with three groups of methods.
Robot-trained ICL models (ICRT~\cite{fu2024icrt},
ICLR~\cite{nguyen2026iclr}, and ContextFlow~\cite{contextflow})
are trained on source tasks and use a target-task
demonstration as context without target-task parameter
updates (Section~\ref{sec:supp_robot_icl}).
Retrieval-based methods (VINN~\cite{pari2022vinn} and
R\&P~\cite{sridhar2025regent}) use the supplied target
demonstration as an action memory with frozen visual
encoders, also without target-task parameter updates
(Section~\ref{sec:supp_retrieval}).
Pretrained robot policies ($\pi_{0.5}$~\cite{black2025pi05}
and GR00T N1.5~\cite{nvidia2025groot15}) are fine-tuned
on source tasks and evaluated with or without further
fine-tuning on the target demonstration
(Section~\ref{sec:supp_pretrained_policies}).

All methods share target-task demonstrations and
evaluation conditions, while their pretraining and
adaptation procedures differ. Our numerical predictor is
not trained on robot data and adapts through numerical
context with its pretrained parameters frozen.

\subsection{Robot-Trained In-Context Models}
\label{sec:supp_robot_icl}

\paragraph{Implementation.}
We use the released implementations of all three methods
and train each on the source data of
Section~\ref{sec:supp_setup}. ICRT uses ICLR's data
pipeline, training recipe, and action interface.

\paragraph{ICRT and ICLR.}
Both train a policy transformer from random initialization
on top of a frozen CrossMAE-RTX ViT-B visual encoder,
pretrained by the ICRT authors on ImageNet and
Open X-Embodiment robot data. ICLR additionally uses its
visual-trace tokens and losses. Each is trained for
125 epochs with AdamW, effective batch size 64, learning
rate $5\times10^{-4}$, weight decay 0.01, and 1.25 warmup
epochs followed by cosine decay, and the final checkpoint
is evaluated. Both predict and execute 16-step action
chunks. On LIBERO, both represent end-effector
orientation internally in the 6D rotation format.
On Meta-World, each training context and query come from
distinct trajectories of the same source scene.

\paragraph{ContextFlow.}
ContextFlow is initialized from the $\pi_0$ base checkpoint
and trained with its released recipe for 20,000 steps at
batch size 32; the final checkpoint is evaluated.
Inference uses ten flow steps and
predicts 50 actions, of which the first five are executed.
On Meta-World, the instruction is the task class name;
the single camera fills the primary image slot with the
wrist view masked, states and actions are padded to the
model interface, and training contexts and
queries are sampled from the same source scene.

\subsection{Retrieval-Based Methods}
\label{sec:supp_retrieval}

\paragraph{VINN.}
VINN uses a ResNet-50 initialized with ImageNet weights and
trained with BYOL on source-task images for 100 epochs;
the encoder, including its batch-normalization statistics,
is frozen during target evaluation. On LIBERO, the two
camera views are encoded separately and their features
concatenated. For each query, VINN retrieves the
$k_{\mathrm{NN}}$ nearest demonstration frames by Euclidean
distance between raw image features, without
proprioception, and averages their actions with weights
proportional to $\exp(-d)$, where $d$ is the feature
distance; the gripper component of the averaged action is
then snapped to the nearest gripper command in the
demonstration.

We select $k_{\mathrm{NN}}$ from $\{1,5,10,20\}$ on 198
source-task context--query pairs, three per LIBERO source
task and per Meta-World source class, drawn from
validation trajectories held out from BYOL training.
The selection minimizes the mean squared error of decoded
single-step actions and gives ten neighbors on both
LIBERO and Meta-World.

\paragraph{R\&P.}
R\&P is the retrieval component of REGENT. It uses the
ImageNet-pretrained ResNet-18 with group normalization
from the REGENT implementation, without further training.
On LIBERO, camera features are concatenated. Each query
retrieves the single nearest demonstration frame by
Euclidean distance between raw image features, without
proprioception, and executes its action.
Both retrieval methods predict and execute one action per
step.

\subsection{Pretrained Robot Policies}
\label{sec:supp_pretrained_policies}

\paragraph{Adaptation protocol.}
Both policies start from their public foundation
checkpoints and are fine-tuned on the source data of
Section~\ref{sec:supp_setup}; the final source checkpoint
is evaluated directly, without a target demonstration.
For +1-shot FT, each one-shot demonstration separately
fine-tunes a copy of this source checkpoint, and the final
fine-tuned checkpoint is evaluated.

\paragraph{$\pi_{0.5}$.}
We use the official openpi implementation. Source
fine-tuning updates all parameters for 30,000 steps with
global batch size 32 and no exponential moving average of
the weights, matching ContextFlow's $\pi_0$
baseline~\cite{contextflow}; the remaining settings follow
openpi's LIBERO recipe for $\pi_{0.5}$, including AdamW
with a peak learning rate of $5\times10^{-5}$ after 10,000
warmup steps. One-shot fine-tuning updates all parameters
for 100 steps with batch size 2 and a constant learning
rate of $2.5\times10^{-6}$, also following ContextFlow.
The policy predicts 10-step action chunks and executes the
first five.

\paragraph{GR00T N1.5.}
We use the official implementation with the GR00T-N1.5-3B
checkpoint. Following its standard fine-tuning setup, the
vision and language backbones remain frozen, while the
projector and diffusion action head are trained. Source
fine-tuning runs 20,000 updates with global batch size 64
and AdamW, using learning rate $10^{-4}$, weight decay
$10^{-5}$, and 5\% linear warmup followed by cosine decay.
One-shot fine-tuning runs 100 updates with batch size 2
and the same optimizer settings. Inference follows the
official LIBERO evaluation example, using eight denoising
steps and predicting 16-step action chunks, of which only
the first action is executed.

\paragraph{Inputs and evaluation.}
Both policies receive the task's language instruction with
the observations of Section~\ref{sec:supp_setup}; on
Meta-World, the instruction is the task class name, and
$\pi_{0.5}$ masks its unused wrist-image slot.
Without target fine-tuning, a policy receives no
demonstration, so each query initialization is evaluated
once, giving 150 rollouts per LIBERO subset and 250 on
Meta-World.

\section{Numerical Predictors and Action Horizons}
\label{sec:supp_predictors}

\subsection{Hyperparameter Selection}
\label{sec:supp_predictor_selection}

\paragraph{Controlled interface.}
The predictor comparison keeps the default numerical
inputs, action targets, context, and horizons ($H=E=8$),
and replaces only TabICLv2. Each conventional predictor is
fitted on the supplied target demonstration, with
hyperparameters selected on source-task validation data
as described below; TabICLv2 keeps its default
configuration.

\paragraph{Validation data.}
We use 198 pairs of complete source-task trajectories.
Each of the 21 LIBERO source tasks contributes three
non-overlapping context--query pairs from six sampled
trajectories, and each of the 45 Meta-World source classes
contributes three pairs, each formed by two trajectories
from one of three sampled scenes. As in target evaluation,
the visual and action projections and the
predictor are fitted on the context trajectory and
evaluated on the query trajectory.

\paragraph{Selection objective.}
We minimize the mean squared error of decoded and clipped
eight-step action chunks, including gripper commands,
averaged over query frames, pairs, tasks, and, for LIBERO,
its three subsets. Each predictor uses one configuration
shared across the LIBERO subsets and a separate
configuration for Meta-World.

\paragraph{Search spaces.}
For kNN, we test $k\in\{1,3,5,10,20\}$ with uniform or
inverse-distance weighting, giving nine candidates because
the two weightings coincide for $k=1$. For Linear, we test
ridge regularization
$\alpha\in\{0.001,0.01,0.1,1,10,100,1000\}$, with
$C=1/\alpha$ for the gripper logistic classifier.
For CatBoost, we test its default configuration and the
nine combinations of depth $\{2,4,6\}$ and learning rate
$\{0.03,0.1,0.3\}$, all with 1,000 iterations.

\paragraph{Selected configurations.}
kNN selects one neighbor on LIBERO and three
inverse-distance-weighted neighbors on Meta-World; Linear
selects $\alpha=100$ and $\alpha=1$; CatBoost selects its
default configuration on LIBERO and depth four with
learning rate 0.03 on Meta-World.

\subsection{Action-Horizon Analysis}
\label{sec:supp_action_horizons}

\begin{table}[htbp]
\centering
\caption{
\textbf{Effect of temporal action formulation.}
We vary the prediction horizon $H$ and execution
horizon $E$ while keeping the frozen visual encoder,
TabICLv2 predictor, demonstrations, and evaluation
protocol unchanged. Results are success rates (\%)
on three LIBERO held-out task subsets.
}
\label{tab:action_horizons_detailed}

\setlength{\tabcolsep}{5pt}
\renewcommand{\arraystretch}{1.12}
\small

\begin{tabular}{cccccc}
\toprule
\multicolumn{2}{c}{\textbf{Action Horizon}}
& \multicolumn{4}{c}{\textbf{LIBERO}} \\
\cmidrule(lr){1-2}
\cmidrule(l){3-6}
Pred. $H$ & Exec. $E$
& Object & Spatial & Goal & Avg. \\
\midrule
1 & 1 & 39.3 & 35.1 & 55.6 & 43.3 \\
4 & 4 & 61.6 & \textbf{43.3} & 68.0 & 57.6 \\
8 & 4 & 64.2 & 39.6 & 74.4 & 59.4 \\
\midrule
\textbf{8} & \textbf{8}
& \textbf{70.2} & 42.2
& \textbf{75.6} & \textbf{62.7} \\
\bottomrule
\end{tabular}
\end{table}

\paragraph{Additional horizon settings.}
Table~\ref{tab:action_horizons_detailed} reports TabICLv2
under further horizon settings, with all other components
fixed. Single-step prediction has no future-motion target;
for $H=4$, 12 PCA components compress the 18 future-motion
values of the next three steps, and $H=8$ uses the default
12 components. Four-step prediction and execution achieves
57.6\% average LIBERO success. With prediction fixed at
$H=8$, executing four actions before observing again
yields 59.4\%, compared with 62.7\% when executing all
eight, isolating the effect of the execution horizon.

\section{Generality Across Model Configurations}
\label{sec:supp_generality}

\begin{figure*}[t]
\centering
\includegraphics[width=\textwidth]{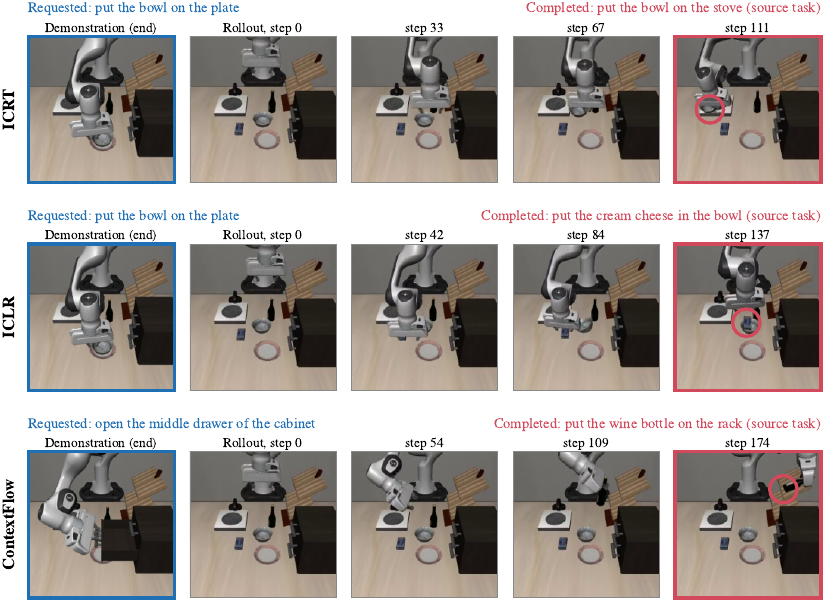}
\caption{\textbf{Examples of robot-trained ICL models completing
source tasks instead of the demonstrated held-out task.}
Each row shows the final frame of the supplied held-out
demonstration and one rollout of the method with that
demonstration; the last frame is taken after a
source-task goal is reached, and circles mark the
moved object.}
\label{fig:source_goal_rollouts}
\end{figure*}

\subsection{Tabular Foundation Models}
\label{sec:supp_tfm_configs}

Each TFM replaces TabICLv2 in the same interface: the
context rows, input features, scalar targets, and action
decoding are identical, and the same frozen model is
queried separately for each scalar target using that
target's context labels. As for TabICLv2, every model uses four
ensemble estimators and the default preprocessing of its
implementation, with no task-specific tuning. Regression
targets take the predictive mean, and gripper targets take
the most probable class. We use
TabPFN-3.5-Fast (84M parameters), Kumo Tabular Small (28M)
and Medium (62M), and Causilo, with separate regression
(37M) and classification (36M) models.

\subsection{Visual Encoders}
\label{sec:supp_encoder_configs}

Each encoder replaces DINOv2-S/14 while all other
settings follow the default configuration, including the
$4\times4$ grid pooling and the visual PCA to at most 64
components per camera. All encoders are frozen and run
under bfloat16 autocast.
SigLIP2-B/16 takes $224\times224$ inputs with its
processor's normalization to $[-1,1]$, and we use the final
hidden states of its vision tower without the
attention-pooling head.
MAE-B/16 uses the original ImageNet-1k MAE checkpoint at
$224\times224$ with ImageNet normalization, taking the
patch tokens after the final normalization.
Both produce $14\times14$ maps of 768-dimensional tokens,
which adaptive average pooling divides into $4\times4$
cells of $4\times4$ tokens, spanning tokens 0--3, 3--6,
7--10, and 10--13 along each axis.
DINOv3-S/16 and DINOv3-B/16 use the LVD-1689M
checkpoints, take the $256\times256$ frame directly with
ImageNet normalization, and produce $16\times16$ maps of
384- and 768-dimensional tokens after the final
normalization, from which the class and register tokens
are discarded. Before PCA, each camera therefore has
6,144 features with DINOv2-S/14 and DINOv3-S/16, and
12,288 with the three base-size encoders.

\subsection{Visual Feature Projection}
\label{sec:supp_visual_projection}

\begin{table}[htbp]
\centering
\caption{\textbf{Effect of visual feature projection.}
One-shot success (\%) on held-out LIBERO tasks with
frozen DINOv2-S/14 and TabICLv2. We vary the projection
and its dimension per camera, keeping the demonstrations,
action targets, and execution protocol fixed.}
\label{tab:visual_projection_detailed}
\setlength{\tabcolsep}{3pt}
\renewcommand{\arraystretch}{1.12}
\footnotesize
\begin{tabular*}{\columnwidth}{@{\extracolsep{\fill}}lcccc@{}}
\toprule
Projection & Object & Spatial & Goal & Avg. \\
\midrule
PCA-16 & \textbf{71.8} & 41.6 & 74.7 & 62.7 \\
PCA-32 & 67.3 & 41.3 & \textbf{80.9} & \textbf{63.2} \\
PCA-64 (default) & 70.2 & \textbf{42.2} & 75.6 & 62.7 \\
Random Projection-64 & 53.6 & 38.9 & 68.2 & 53.6 \\
\bottomrule
\end{tabular*}
\end{table}

Table~\ref{tab:visual_projection_detailed} compares visual
PCA dimensions with a random projection of equal dimension.
The random projection replaces the PCA directions with a
fixed, data-independent orthogonal basis per camera,
keeping the same demonstration-mean centering.
PCA with 16, 32, and 64 components per camera yields
similar average LIBERO success (62.7\%, 63.2\%, and
62.7\%), although individual subsets vary. Replacing
PCA-64 with the random projection reduces the average to
53.6\%. Within the tested range, the choice of projection
directions has a larger effect on average success than the
PCA dimension.

\section{Controlled Goal Switching}
\label{sec:supp_goal_switching}

\paragraph{Protocol.}
The goal-switching experiment uses held-out LIBERO-Goal
tasks 0, 4, and 8, whose 20 common initial simulator states
produce identical initial images across the three task
environments. Unlike the main evaluation, rollouts do not
terminate on success; they run for 300 environment steps
while all ten LIBERO-Goal predicates are recorded.
A rollout achieves the requested goal if its predicate
holds at any step. The source-demonstration control
evaluates the robot-trained ICL methods with one
demonstration of each of the seven source tasks from the
same 20 initial states, giving 140 rollouts per method.

\begin{table}[htbp]
\centering
\caption{
\textbf{Goal switching and source-task controls under identical
initial observations.}
\emph{Requested}: success rate (\%) on three held-out
Goal tasks. \emph{Source goal}: number of the same 180
rollouts that complete a source-task goal without
completing the requested or another held-out goal.
\emph{Source demo}: success rate (\%) over 140 rollouts
using demonstrations of the seven source tasks and
the same initial states. Dashes indicate that this
control was not evaluated for that method.
}
\label{tab:goal_switch_diagnostics}

\setlength{\tabcolsep}{4pt}
\renewcommand{\arraystretch}{1.12}
\small

\begin{tabular*}{\columnwidth}{
@{\extracolsep{\fill}}lccc@{}}
\toprule
\multirow[c]{2}{*}[-0.4ex]{\textbf{Method}}
& \multicolumn{2}{c}{\textbf{Held-out demo}}
& \textbf{Source demo} \\
\cmidrule(lr){2-3}
\cmidrule(l){4-4}

& Requested $\uparrow$
& Source goal $\downarrow$
& Success \\
\midrule
ICRT~\cite{fu2024icrt}         & 0.6 & 27 & 87.1 \\
ICLR~\cite{nguyen2026iclr}     & 0.0 & 5  & 85.7 \\
ContextFlow~\cite{contextflow} & 4.4 & 87 & 90.0 \\
\midrule
VINN~\cite{pari2022vinn} & 7.2 & \textbf{0} & -- \\
R\&P~\cite{sridhar2025regent} & 48.3 & \textbf{0} & -- \\
\midrule
\textbf{\ours{}}
& \textbf{75.6} & \textbf{0} & -- \\
\bottomrule
\end{tabular*}
\end{table}

\paragraph{Source-task controls.}
\label{sec:supp_source_controls}
The robot-trained ICL models succeed on their source tasks:
with source demonstrations and the same initial states,
ICRT, ICLR, and ContextFlow reach 87.1\%, 85.7\%, and
90.0\%, respectively
(Table~\ref{tab:goal_switch_diagnostics}). Under held-out
demonstrations, 27, 5, and 87 of their 180 rollouts instead
complete source-task goals, accounting for part of their
failures to follow unseen goals.
For example, given a demonstration of putting the bowl on
the plate, ICRT puts the bowl on the stove, and given a
demonstration of opening the middle drawer, ContextFlow
puts the wine bottle on the rack, both of which are source
tasks (Fig.~\ref{fig:source_goal_rollouts}).
The original ContextFlow evaluation shows the same gap
under its own task split: ContextFlow and ICRT reach
90.0\% success on seen LIBERO-Goal tasks but 0.0\% and
1.0\% on held-out ones~\cite{contextflow}.

\section{Multi-Output Prediction for Lower Latency}
\label{sec:efficiency}

This exploratory extension targets deployment cost.
Our default TabICLv2 implementation predicts each scalar
action target separately, as is standard for current TFMs,
so latency grows with the number of targets per chunk.
We pretrain Lattice-MT, a lightweight TFM, on
generic numerical tasks, without robot trajectories, to
predict groups of up to eight targets jointly, and freeze
it for robot evaluation. The model will be released at
\url{https://huggingface.co/hachure-ai/Lattice-MT}.

\begin{table}[htbp]
\centering
\caption{\textbf{Multi-output prediction.}
One-shot success (\%) with the same numerical inputs,
targets, demonstrations, and evaluation protocol;
only the frozen predictor changes. MW: Meta-World. Latency is the
median time to generate an eight-step chunk, and
throughput is amortized over executed actions, both at
batch size one.}
\label{tab:multi_output}
\setlength{\tabcolsep}{2pt}
\renewcommand{\arraystretch}{1.12}
\footnotesize
\begin{tabular*}{\columnwidth}{@{\extracolsep{\fill}}lccccccc@{}}
\toprule
\multirow[c]{2}{*}[-0.4ex]{\textbf{Predictor}}
& \multicolumn{4}{c}{\textbf{LIBERO}}
& \multirow[c]{2}{*}[-0.4ex]{\textbf{MW}}
& \multirow[c]{2}{*}[-0.4ex]{\shortstack[c]{\textbf{Lat.}\\(ms)}}
& \multirow[c]{2}{*}[-0.4ex]{\shortstack[c]{\textbf{Thr.}\\(Hz)}} \\
\cmidrule(lr){2-5}
& Obj. & Spa. & Goal & Avg. & & & \\
\midrule
TabICLv2 & 70.2 & 42.2 & \textbf{75.6} & 62.7 & \textbf{49.3} & 881 & 9.6 \\
Lattice-MT & \textbf{73.6} & \textbf{44.4} & 72.7 & \textbf{63.6} & 47.7 & \textbf{229} & \textbf{34.2} \\
\bottomrule
\end{tabular*}
\end{table}

We measure latency on an RTX 4090 with batch size one,
from receiving an observation to returning executable
actions. Timing covers visual encoding, PCA
projection, TFM prediction, and action decoding, and
excludes simulation and the one-time context preparation.
Chunk latency is the median time to generate a new
eight-step chunk; amortized throughput divides the number
of executed actions by the total policy time, including
steps that execute queued actions.
Lattice-MT reduces chunk latency from 881\,ms
to 229\,ms (3.8$\times$), raises amortized throughput from
9.6 to 34.2\,Hz, and achieves comparable average LIBERO
success (63.6\% versus 62.7\%) and Meta-World success
(47.7\% versus 49.3\%; Table~\ref{tab:multi_output}).
These results suggest that the latency of \ours{} stems
largely from per-target prediction in current TFMs rather
than from the numerical formulation itself. Because \ours{}
uses a frozen predictor and requires no robot-specific
training, faster TFMs can replace TabICLv2 directly, and the
latency of \ours{} can be expected to decrease as TFMs
continue to develop.

\end{document}